\documentclass{article}

\usepackage{times}
\usepackage{graphicx}
\usepackage{amsmath}
\usepackage{amssymb}
\usepackage{booktabs}
\usepackage{tabularx,ragged2e}
\usepackage{array}
\usepackage{color, colortbl}
\usepackage{amsfonts}       
\usepackage{nicefrac}       
\usepackage{microtype}      
\usepackage{epsfig}
\usepackage{amsmath}
\usepackage{amssymb}
\usepackage{xcolor}
\usepackage{mathtools}
\usepackage{lipsum}
\usepackage{booktabs}       
\usepackage{wrapfig}
\usepackage{caption}
\usepackage{subcaption}
\usepackage[ruled,vlined]{algorithm2e}
\usepackage{stfloats}
\usepackage{bold-extra}
\usepackage{fancyvrb}
\usepackage{amsthm}
\usepackage{onimage}

\usepackage{geometry}
\usepackage[numbers]{natbib}
\usepackage{fullpage}
\date{}

\definecolor{LightCyan}{rgb}{0.8,1,1}
\definecolor{LightGray}{rgb}{0.83, 0.83, 0.83}
\definecolor{LightYellow}{rgb}{1.0, 1.0, 0.88}
\definecolor{MistyRose}{rgb}{1.0, 0.89, 0.88}
\newcolumntype{Y}{>{\centering\arraybackslash}X}

\newcolumntype{s}{>{\centering\arraybackslash\hsize=.6\hsize}X}
\newcolumntype{m}{>{\centering\arraybackslash\hsize=1.3\hsize}X}

\usepackage[pagebackref=true,breaklinks=true,colorlinks,bookmarks=false]{hyperref}

\begin{document}


\title{Tilted Cross-Entropy (TCE): Promoting Fairness in Semantic Segmentation}


\author{Attila Szab\'{o}\textsuperscript{\rm 1}\thanks{ Equal contributions.}\quad~Hadi Jamali-Rad\textsuperscript{\rm 1, 2}\footnotemark[1]\quad~Siva-Datta Mannava\textsuperscript{\rm 1}\\
		{\textsuperscript{\rm 1}Shell Global Solutions International B.V., Amsterdam, The Netherlands}  \\
		{\textsuperscript{\rm 2}Delft University of Technology (TU Delft), Delft, The Netherlands} \\
		\small{\texttt{\{attila.szabo,hadi.jamali-rad,siva-datta.mannava\}@shell.com}},\\ \small{\texttt{h.jamalirad@tudelft.nl}}
	}



\maketitle


\begin{abstract}
\vspace{-0.3cm}
   Traditional empirical risk minimization (ERM) for semantic segmentation can disproportionately advantage or disadvantage certain target classes in favor of an (unfair but) improved overall performance. Inspired by the recently introduced tilted ERM (TERM), we propose tilted cross-entropy (TCE) loss and adapt it to the semantic segmentation setting to minimize performance disparity among target classes and promote fairness. Through quantitative and qualitative performance analyses, we demonstrate that the proposed Stochastic TCE for semantic segmentation can efficiently improve the low-performing classes of Cityscapes and ADE20k datasets trained with multi-class cross-entropy (MCCE), and also results in improved overall fairness. \vspace{-0.4cm}
\end{abstract}

\section{Introduction and Related Work}
\label{sec:intro}
\vspace{-0.3cm}
Semantic segmentation is a fundamental problem in computer vision and a pivotal step towards content-based image analysis and scene understanding. It has received an upsurge of attention recently owing to its wide variety of applications in medical imaging \cite{ronneberger2015u, rezaei2017conditional}, autonomous driving \cite{menze2015object, cordts2016cityscapes}, satellite image processing \cite{volpi2015semantic, henry2018road}, and robotics \cite{geiger2013vision, shvets2018automatic}, to name a few. Early segmentation methodologies are mostly developed with clustering algorithms at their core \cite{kass1988snakes, nock2004statistical, plath2009multi, minaee2019admm}. Recent advances in deep learning have revolutionized this field resulting in state-of-the-art (SoTA) image segmentation algorithms such as FCN \cite{long2015fully}, U-Net \cite{ronneberger2015u}, PSPNet \cite{zhao2017pyramid}, EncNet \cite{zhang2018context}, Exfuse \cite{zhang2018exfuse}, DeepLabv3+ \cite{chen2018encoder}, PS and Panoptic DeepLab \cite{kirillov2019panoptic, cheng2019panoptic}, HRNet \cite{wang2020deep} and many other elegant architectures that considerably outperformed the traditional signal processing-based methods.

The choice of loss function is essential for applicability of semantic segmentation in different contexts. Therefore, several studies have investigated the impact of modified loss functions on the performance of semantic segmentation models in generic and use-case related datasets. A comprehensive list of some of these recent loss functions is provided in four different categories in \cite{jadon2020survey}: i) distribution-based losses such as binary class cross-entropy (BCE), ii) region-based losses such as Dice loss \cite{sudre2017generalised} and its variants, iii) boundary based losses such as Hausdorff distance \cite{ribera2018weighted, karimi2019reducing}, and iv) compound losses such Exponential Logarithmic loss \cite{wong20183d}. Generic empirical risk minimization (ERM) loss functions such as BCE or Dice can disproportionately advantage or disadvantage some classes in favors of an improved average performance. This results in models that treat certain classes (e.g. with less presence in the training dataset) in an \emph{unfair} fashion. Somewhat related to this concern, significant research has been conducted on proposing alternatives to the cross-entropy (CE) loss for semantic segmentation to handle imbalanced data using weighted cross entropy \cite{pihur2007weighted}, balanced cross entropy \cite{xie2015holistically}, and the Focal loss \cite{lin2017focal}. However, none of these approaches directly address the \emph{fairness} problem. Being the most relevant approach among the mentioned solutions, Focal loss \cite{lin2017focal} down-weights the contribution of easy examples and enables the model to focus more on learning hard ones. In a broader scope of optimization for machine learning, \cite{hashimoto2018fairness} proposes a solution to ensure different subgroups within a population are treated fairly and \cite{duchi2016variance} develops a solution with favorable out-of-sample performance. Most recently, a unified framework called tilted ERM (TERM) has been proposed in \cite{li2020tilted} to flexibly address the deficiencies of traditional ERM with respect to handling outliers and treating subgroups fairly. It is demonstrated in \cite{li2020tilted} that TERM can not only efficiently promote fairness in a multitude of applications, but also outperforms the likes of Focal Loss \cite{lin2017focal} and RobustRegRisk \cite{duchi2016variance} in different settings. Besides, \cite{li2020tilted} provides efficient batch and stochastic first-order optimization methods for solving TERM. Inspired by flexibility and superior performance of TERM in promoting fairness, we have studied the impact of adopting a similar approach in a new context, i.e., semantic segmentation. 

Our contributions are as follows: i) We propose to employ tilted cross-entropy (TCE) as a novel loss to promote fairness in semantic segmentation. ii) We adapt the derivations of \cite{li2020tilted} to fit into semantic segmentation setting and reformulate the commonly used CE loss as TCE. We then propose a stochastic non-hierarchical optimization algorithm for solving TCE. iii) We empirically demonstrate the effectiveness of Stochastic TCE on promoting fairness in semantic segmentation for Cityscapes (and ADE20k in appendix). \footnote{The code will be publicly available soon.} 

\section{Tilted CE (TCE) for Semantic Segmentation}
\label{sec:term_semseg}
\vspace{-0.3cm}
In this section, we introduce TCE: an adapted and reformulated version of TERM \cite{li2020tilted} for semantic segmentation. We then propose a stochastic non-hierarchical batch solution of TCE. In the following, we denote the cardinality of $\mathcal{X}$ as $|\mathcal{X}|$, and the set $\{1, \cdots, n\}$ as $[n]$. Let $\mathcal{D}_t =\{({\bf X},{\bf Y})_{1},...,({\bf X},{\bf Y})_{M}\}$ be the training dataset containing $M$ samples with ${\bf X}_m$ and ${\bf Y}_m$ respectively denoting the $m$th image and its corresponding label map (also called mask). Here, ${\bf X}$ is of size $H \times W \times 3$ for RGB images with a total of $H \times W = N$ pixels. The corresponding label map ${\bf Y}$ is of size $H \times W$ with elements in $[K]$. ${\bf X}_m$ contains a maximum of $K$ classes each occupying $n_{m, c}$ pixels, where $\sum_{c \in [K]} n_{m, c} = H \times W = N, \forall m$. The most commonly used ERM loss for semantic segmentation is the pixel-wise loss $\mathcal{L} = \sum_{m=1}^M\textup{CE}({\bf Y}_m, \hat{{\bf Y}}_m)$ \cite{chen2018encoder, xiao2018unified, wang2020deep}, which is computed using a multi-class CE between the $1$-hot encoded versions of the original label map ${\bf Y}$ and the inferred one $\hat{{\bf Y}}$:
\begin{equation}
\label{eq:erm}
    \mathcal{L} = - \frac{1}{MN}\sum_{m=1}^M \sum_{c = 1}^K \sum_{i = 1}^{n_c}  y_{m,c,i}\, \log(\hat{y}_{m,c,i}),
\end{equation}
where $y_{m,c,i}$ denotes the $i$th pixel in the $c$th class of ${\bf Y}_m$. \cite{li2020tilted} proposes to \emph{tilt} ERM $\mathcal{R}(\theta) = 1/M \sum_{i \in [M]} f(x_i, \theta)$ as $\tilde{\mathcal{R}}(\theta, t) = 1/t \log(1/M \sum_{i \in [M]} e^{tf(x_i, \theta)})$, with loss function $f(x_i, \theta)$ and model parameters $\theta$. The tilt parameter $t$ can be tuned to flexibly promote robustness or fairness. In theory, setting $t=0$ recovers ERM (i.e, $\tilde{\mathcal{R}}(\theta, 0) = \mathcal{R}(\theta)$), and as $t \rightarrow \infty$, TERM minimizes the worst loss, thus ensuring the model is a reasonable fit for all samples \cite{li2020tilted}. With this in mind, there are (at least) two levels around which a sensible tilting of multi-class CE (MCCE) \eqref{eq:erm} can be implemented. More specifically, we can tilt \eqref{eq:erm} at i) image (or sample) level and ii) class level. Let us start with the image level. To tilt at \emph{image level}, we need to reformulate \eqref{eq:erm} as    
\begin{align}
\label{eq:term_m}
    \begin{split}
        \tilde{\mathcal{L}}^{img} &= \frac{1}{t} \log\Big( \frac{1}{M}\sum_{m=1}^M e^{t\, \mathcal{L}_m} \Big),\\
        \textup{where:} \quad \mathcal{L}_m &= - \frac{1}{N} \sum_{c = 1}^K \sum_{i = 1}^{n_c}  y_{m,c,i}\, \log(\hat{y}_{m,c,i}).
    \end{split}
\end{align}
Following the same strategy, to tilt at \emph{class level} per image we need to reformulate \eqref{eq:erm} as 
\begin{align}
\label{eq:term_c}
    \begin{split}
        \tilde{\mathcal{L}}^{cls} &= \frac{1}{M}\sum_{m=1}^M \frac{1}{t} \log\Big( \frac{1}{K}\sum_{c = 1}^K e^{t\, \mathcal{L}_{m,c}} \Big),\\
        \textup{where:} \quad \mathcal{L}_{m,c} &= - \frac{1}{n_c}\sum_{i = 1}^{n_c}  y_{m,c,i}\, \log(\hat{y}_{m,c,i}).
    \end{split}
\end{align}

\subsection{Solving TCE for Semantic Segmentation}
\label{ssec:stoachstic}
\vspace{-0.3cm}
Depending on the tilt level, one has to replace the pixel-wise MCCE part of the semantic segmentation loss with one of the proposed losses in \eqref{eq:term_m} and \eqref{eq:term_c}. In our experience, directly plugging these loss functions into semantic segmentation optimization problem could lead to convergence issues and caution has to be put in place. An alternative approach to solve TCE (applicable to both sample and hierarchical levels) is to follow along the \emph{stochastic} approach proposed in \cite{li2020tilted}. It is proven in \cite{li2020tilted} that the gradient of the tilted loss $\tilde{\mathcal{L}}$ is a weighted average of the gradients of the original individual losses, where each data point is weighted exponentially proportional to the value of its loss. This is the key idea behind the proposed dynamic weight updating and sampling strategy of Stochastic TCE laid out in Algorithm~\ref{alg:sotchastc_term}. Let us dive deeper and walk through the algorithm. For the sake of simplicity, here we drop the superscript denoting the tilting level of $\tilde{\mathcal{L}}$ in \eqref{eq:term_m} and \eqref{eq:term_c}, and use a subscript to refer to the class $\tilde{\mathcal{L}}_c$ and batch of data within the class $\tilde{\mathcal{L}}_B$. The class weights $w_c$ are stored/updated in $W$.
\LinesNumbered
\begin{algorithm}[t!]
	\SetKwInput{Require}{Require}
	\SetKwInput{Initialize}{Initialize}
	\SetAlgoLined
	\DontPrintSemicolon
	\SetNoFillComment
	\Initialize{$w_c = \tilde{\mathcal{L}}_c = 0, \forall c \in [C]$, $\theta$}
	\Require{$\gamma$, $\eta$, $t$}
	Divide $\mathcal{D}^t$ into $\mathcal{D}^t_c, \forall c \in [C]$\; 
	\While{\textup{stopping criteria not met}}{
		sample class $c \in [C]$ from a categorical distribution with probabilities $w_c \in W$\;  
		sample minibatch $B$ within $\mathcal{D}^t_c$\;
		$\mathcal{L}_B \gets$ compute the loss \eqref{eq:erm} on $B$\;
		tilt the batch loss: $\tilde{\mathcal{L}}_B \gets e^{t \mathcal{L}_B}$\;
		$\tilde{\mathcal{L}}_c \gets (1 - \gamma)\,\tilde{\mathcal{L}_c} + \gamma \tilde{\mathcal{L}}_B$\; 
		$w_c \gets \tilde{\mathcal{L}}_c / (\sum_{l=1}^K \tilde{\mathcal{L}}_l), \forall c \in [C]$\;
		Update model parameters: $\theta \gets \theta - \nabla \mathcal{L}_B$
		}
	\caption{Stochastic TCE for Segmentation}\label{alg:sotchastc_term}
\end{algorithm}

The algorithm starts by dividing the traninig dataset $\mathcal{D}^t$ into $C$ subsets $\mathcal{D}^t_c$ each containing the images corresponding to individual classes. Note that an image can contain multiple classes, and thus, these sets can overlap. One can also consider forming non-overlapping sets based on $\mathcal{D}^t$. Per propagation round, one class (let us say $c$) will be selected from the categorical distribution $[C]$ with probabilities (weights) $W$ (line $3$). These weights are dynamically updated (line $8$). Next, a minibatch $B$ is sampled from the training data of the selected class $\mathcal{D}^t_c$ and the tilted batch loss $\tilde{\mathcal{L}}_B$ is calculated on $B$ (lines $5$ and $6$). Line $7$ proposes a linear dynamic with rate $\gamma$ to update the tilted loss of the selected class $\tilde{\mathcal{L}}_c$ based on its previous value and the current batch estimate $\tilde{\mathcal{L}}_B$. The weight of class $c$, $w_c$, will then be updated using a normalization applied to all the tilted losses (line $8$). These dynamically updated weights, $w_c \in W$, will be used in the next iteration to decide from which class to sample. Finally, model parameters in $\theta$ are updated.

\section{Experimental Setup}
\label{sec:exp}
\vspace{-0.3cm}
Here, we assess the impact of TCE on one of the most commonly adopted datasets for semantic segmentation, Cityscapes \cite{cordts2016cityscapes}. The evaluation results for ADE20k \cite{zhou2019semantic} can be found in the appendix. Cityscapes contains $2,975$ train and $500$ validation images from $19$ main target classes. 

%
\begin{table*}[t!]
    \footnotesize
	\caption{Performance comparison on Cityscapes \emph{validation} set, sorted based on DLv3+ with MCCE.}
	\vspace{-0.35cm}
	\label{tb:cityscapes_val}
	\centering
	{\tabcolsep=0pt\def\arraystretch{1.0}
	\begin{tabularx}{400pt}{@{}l Y Y Y Y Y Y Y Y Y Y  Y Y@{}}
		\toprule
		Method     	& wall & train & rider & fence & terrain & truck & m.cycle & pole & bus & t. light \\
		\midrule
         MCCE \cite{chen2018encoder}  & 48.46 & 53.66 & 61.12 & {\bf62.23} & 62.98 & 66.21 & 67.72 & 69.19 & 71.53 & 74.33 \\
        Focal loss \cite{lin2017focal} & 49.11 & {\bf79.55} & {\bf67.56} & 60.75 & 62.31 & {\bf73.16} & 67.71 & 64.53 & {\bf85.02} & 68.92 \\ 
		\rowcolor{LightCyan}
		TCE \tiny{$t=.1$} &49.36 & 76.75 & 66.64 & 60.38 & {\bf65.69} & 72.03 & 69.09 & 69.34 & 75.61 & 73.64 \\
		\rowcolor{LightYellow}
		TCE \tiny{$t=1$} & {\bf53.47} & {\bf79.32} & 65.67 & 59.25 & 63.74 & 64.32 & {\bf69.45} & {\bf69.54} & 66.05 & {\bf74.51} \\
		\toprule
		\toprule
		continued	& bicycle & t. sign & person & sidewalk & vegetation & building & sky & car & road	&\multicolumn{1}{c}{\cellcolor{gray!35}mIoU}\\
		\midrule
		MCCE \cite{chen2018encoder} & 79.23 & {\bf81.22} & 83.19 & 84.95 & {\bf92.52} & {93.04} & 95.00 & {\bf95.45} & 98.06 & \multicolumn{1}{c}{\cellcolor{gray!35}75.79} \\
		Focal loss \cite{lin2017focal} & 77.71 & 77.62 & 81.23 & 81.56 & 91.54 & 91.89 & 93.71 & 94.67 & 97.14 & \multicolumn{1}{c}{\cellcolor{gray!35}77.14} \\
		\rowcolor{LightCyan}
		TCE \tiny{$t=.1$} &{\bf79.89} & 81.77 & {\bf84.00} & {\bf86.24} & 92.43 & {\bf93.16} & {\bf95.34} & {\bf95.47} & {\bf98.27} & \multicolumn{1}{c}{\cellcolor{gray!35}\bf78.16} \\
		\rowcolor{LightYellow}
		TCE \tiny{$t=1$} & {\bf79.92} & 81.04 &{\bf83.96} & {\bf86.22} & 92.44 & 92.92 & 95.11 & 94.53 & {\bf98.28} &\multicolumn{1}{c}{\cellcolor{gray!35}77.35} \\
		\bottomrule
	\end{tabularx}}
	\vspace{-0.25cm}
\end{table*}
%

%
\begin{table*}[t]
    \footnotesize
	\caption{Performance comparison on Cityscapes \emph{validation} set, sorted based on SoTA DLv3+.}
	\vspace{-0.35cm}
	\label{tb:cityscapes_val_sota}
	\centering
	{\tabcolsep=0pt\def\arraystretch{1.0}
	\begin{tabularx}{400pt}{@{}l Y Y Y Y Y Y Y Y Y Y Y Y@{}}
		\toprule
		Method     	& wall &fence	&rider	&terrain	&m.cycle	&pole	&t. light	&bicycle	&t. sign	&train \\
		\midrule
        SoTA MCCE \cite{chen2018encoder} & {\bf57}.26 &{\bf62.18} &62.76 &63.38 & 64.50 &65.11 &68.41	 &77.26	 &78.78 &{\bf80.90}\\
		\rowcolor{LightCyan}
		TCE \tiny{$t=.1$} &49.36 & 60.38 & {\bf66.64} & {\bf65.69} & {\bf69.09} & {\bf69.34} & {\bf73.64} & {\bf79.89} & {\bf81.77 }& 76.75		\\ 
		\toprule
		\toprule
		continued	& person	&sidewalk	&truck	&bus	&vegetation	&building	&sky	&car	&road	&\multicolumn{1}{c}{\cellcolor{gray!35}mIoU}\\
		\midrule
		SoTA MCCE \cite{chen2018encoder}  & 82.14	&84.7	&{\bf85.31}	&{\bf89.07}	&{\bf92.65}	&92.69	&95.29	&{95.31}	&98.13 &\multicolumn{1}{c}{\cellcolor{gray!35}\bf78.73} \\
		\rowcolor{LightCyan}
		TCE \tiny{$t=.1$}  & {\bf84.00} & {\bf86.24} & 72.03 & 75.61 & 92.43 & {\bf93.16} & {\bf95.34} &{\bf 95.47} & {\bf98.27} & \multicolumn{1}{c}{\cellcolor{gray!35}78.16}\\
		\bottomrule
	\end{tabularx}}
	\vspace{-0.1cm}
\end{table*}

\textbf{Training strategy and baselines.} Our trainings are run separately on standard Microsoft Azure $4$-GPU P100 Tesla nodes, each with $16$GB of memory. For experiments on Cityscapes, we used DeepLabv3+ (also referred to as DLv3+) \cite{chen2018encoder} with ReseNet-101 backbone as our reference implementation of multi-class CE (MCCE), and on top of that we have implemented TCE. DLv3+ is among the top performing model architectures for Cityscapes. Following \cite{chen2018encoder}, we used minibatch SGD with learning rate $l_r = 0.01$ and momentum $0.9$ for all models, and adjusted for a total minibtch size of $8$ ($2$ per GPU). The reported results of \cite{chen2018encoder} are based on our own trainings, for the sake of a fair comparison. Image crop size and other pre/post-processing parameters are set per default as suggested in \cite{chen2018encoder}. We also compare our performance against the Focal loss for semantic segmentation \cite{lin2017focal} with the best parameters $\gamma = 2$ and (class weights) $\alpha$ set to the inverse (normalized) class pixel counts computed across the whole dataset. 

\textbf{Fairness and its evaluation criteria.} The notion of fairness in this setting is promoting a more consistent (and less varied) performance across different classes. This is to ensure that there are less (or ideally no) classes that have been significantly disadvantaged, due to for instance less presence or difficult characteristic features, for the sake of a higher average performance. Promoting fairness by minimizing performance disparity is also a core idea of TERM \cite{li2020tilted}, and resonates with other recent approaches to fairness \cite{hashimoto2018fairness, li2019fair, mohri2019agnostic}. More concretely, i) best worst-case performance \cite{hashimoto2018fairness, mohri2019agnostic}, and ii) least variance across clients/classes \cite{li2019fair} are recently proposed to promote/evaluate fairness across a set of tasks or networked clients. We investigate both measures as our key criteria. More specifically, besides the overall mean-intersection-over-union (mIoU), we compare the models on: i) sorted (w.r.t MCCE) bottom and top $25\%$ mIoUs; ii) bottom and top $25$th percentiles; and iii) standard deviation and worst case performance (in IoU) across classes. Note that the overall mIoU does not have to be improved when applying TCE; the goal is to minimize performance disparity which can sometimes come at the cost of lower overall mIoU.   

\section{Evaluation Results}
\label{sec:eval_res}
\vspace{-0.3cm}
\begin{figure*}[t!]
	\centering
    \includegraphics[trim={0cm 3.15cm 0cm 0cm},clip,width=0.99\textwidth]{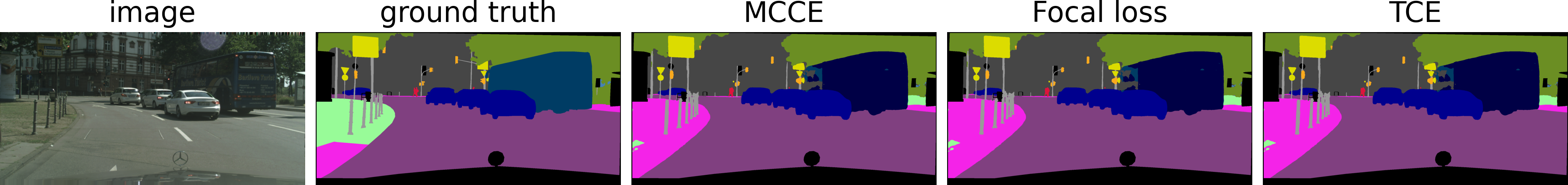}
	\begin{tikzonimage}[width=0.99\textwidth]{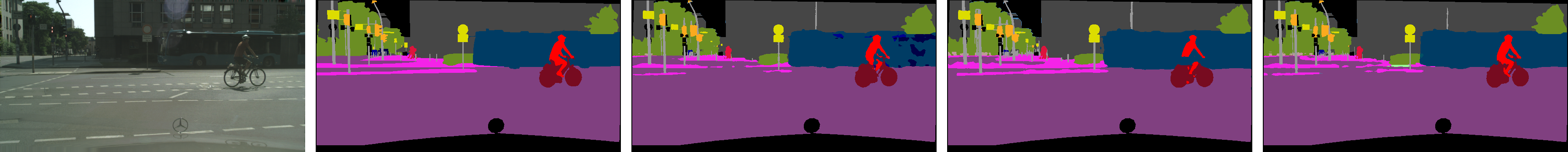}
		\draw[yellow, very thick, rounded corners, dashed] (0.9,0.44) rectangle (0.999,0.86);
		\draw[yellow, very thick, rounded corners, dashed] (0.945,0.395) rectangle (0.981,0.79);
	\end{tikzonimage}
	\begin{tikzonimage}[width=0.99\textwidth]{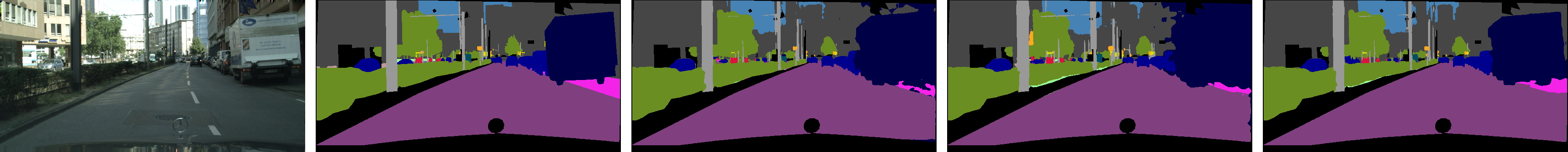}
		\draw[yellow, very thick, rounded corners, dashed] (0.945,0.37) rectangle (0.999,0.995);
	\end{tikzonimage}
	\begin{tikzonimage}[width=0.99\textwidth]{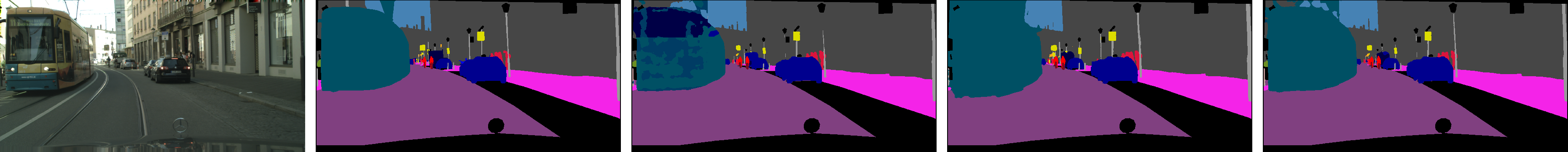}
		\draw[yellow, very thick, rounded corners, dashed] (0.805,0.36) rectangle (0.865,0.972);
	\end{tikzonimage}
	\begin{tikzonimage}[width=0.99\textwidth]{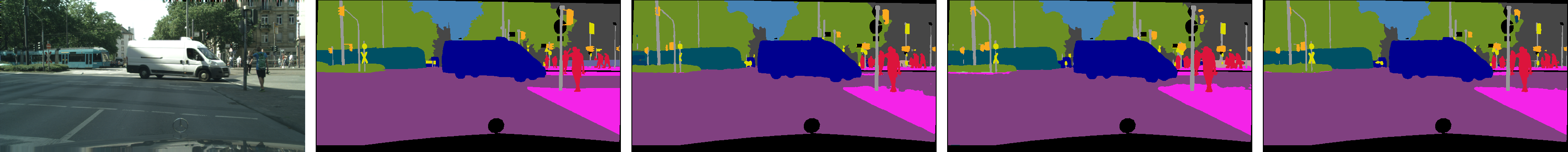}
		\draw[yellow, very thick, rounded corners, dashed] (0.805,0.465) rectangle (0.879,0.75);
		\draw[yellow, very thick, rounded corners, dashed] (0.928,0.34) rectangle (0.999,0.485);
	\end{tikzonimage}
	\caption{Impact of TCE on improving low-performing classes of MCCE. Best view in color with $300\%$ zoom.}
	\vspace{-0.4cm}
	\label{fig:comparison_1}
\end{figure*}

Table~\ref{tb:cityscapes_val} compares the sorted mIoU breakdown of DLv3+ with ResNet-101 backbone \cite{chen2018encoder} trained with the standard multi-class cross-entropy (MCCE) against the same model retrained with the Focal loss and the proposed TCE for $t = 0.1$ and $1$. Even though we are not \emph{necessarily} expecting an improvement in the overall mIoU, here for all $t$'s, the overall performance with TCE has improved beyond same model trained with MCCE ($+2\%$ for $t = 0.1$) and Focal loss (about $1\%$ for $t = 0.1$). TCE with $t = 1$ is expected to push more towards minimizing performance disparity and thus promoting fairness (as is also shown in Table~\ref{tb:fairness_focal_Cityscapes}), and it shows the best improvement in the least performing classes such as ``wall'' and ``train''. The same model architecture, DeepLabv3+ (abbreviated to DLv3+ here), reports better performance results with a more complex backbone Xception-65 \cite{chen2018encoder}. We could not reproduce those results because we did not have access to large enough nodes on Microsoft Azure to accommodate this backbone with batch sizes larger than $8$. Nonetheless, we were curious to know how TCE implemented on top of a model with a weaker backbone would compare with the the state-of-the-art (SoTA) results reported in \cite{chen2018encoder}. This comparison is summarized in Tables~\ref{tb:cityscapes_val_sota}. Interestingly, even compared to the SoTA model with improved backbones, TCE is still improving on several low-performing classes (such as ``rider", ``terrain'', ``motorcycle'', etc.)  in favor of promoting fairness across target classes. 

%
%
\begin{table}[t!]
    \footnotesize
	\caption{Fairness criteria for Cityscapes \cite{cordts2016cityscapes}}
	\vspace{-0.35cm}
	\label{tb:fairness_focal_Cityscapes}
	\centering
	{\tabcolsep=0pt\def\arraystretch{1.0}
	
		\begin{tabularx}{230pt}{l  s  s | m m | s s }
			\toprule
			& \multicolumn{2}{c}{sorted $25\%$ } & \multicolumn{2}{c}{$(25^{th} \textup{perc., mIoU})$}  & \multicolumn{2}{c}{overall}
		    \tabularnewline
			\cmidrule(lr){2-3}\cmidrule(lr){4-5}\cmidrule(lr){6-7}
			Method   & bottom     & top        & bottom    &top  & worst & std. 				\tabularnewline 
			\midrule
			MCCE \cite{chen2018encoder} & 57.69 & {94.81} & (64.60, 57.69) & (88.74, {94.81}) & 48.46 & 14.96  \\
			Focal loss \cite{lin2017focal} & 63.86 & 93.79 & (67.64, 60.85) &(88.28, 93.79) & 49.11 & {\bf13.35} \\
			\rowcolor{LightCyan}
			TCE \tiny{$t=.1$}& 63.76 & 94.93 & (69.22, \bf{62.23}) &(89.34, 94.93) & 49.36 & {\bf13.34} \\
			\rowcolor{LightYellow}
			TCE \tiny{$t=1$} & {\bf64.29} & 94.66 & (65.86, {61.29}) &(89.33, 94.66)& {\bf53.47} &13.57 \\
			\bottomrule
	\end{tabularx}}
\end{table} 

These two tables illustrate promising improvement in low-performing classes the extent of which is further investigated and is summarized in Table~\ref{tb:fairness_focal_Cityscapes}. Here, the following three measures are presented. First, the sorted bottom and top $25\%$ mIoU. To compute this, the target classes are sorted based on the IoU performance breakdown of the the model trained with MCCE. Then for each model the mIoU of the bottom and top $25\%$ ($5$ classes out of $19$) are taken into account. This is to demonstrate the impact of TCE compared to the reference MCCE. Here, improved mIoU for bottom classes (even at cost the of lower mIoU for the top ones) indicates more fairness. Second, in a tuple, the IoU threshold corresponding to (bottom and top) $25$th percentile of each model and the mIoU of the classes falling within the percentile are presented. Note that in this case each model will be sorted according to its own target class IoU's. The idea is that improvement in bottom percentile threshold and corresponding mIoU could be indicative of improved fairness. Again, we do not expect improvement but potential drop in the top percentiles; they are reported to provide a more complete picture. Third, the overall fairness measures \cite{hashimoto2018fairness, li2019fair, mohri2019agnostic}, i.e., the worst performance among target classes and the standard deviation across class IoU's (denoted respectively as worst and std. in the table) are presented. 

The results show that on all three metrics TCE offers improved fairness when compared to MCCE and Focal loss. Let us focus on $t = 1$. In sorted bottom $25\%$, we gain $+6\%$ and $+0.4\%$ beyond MCCE and Focal loss, respectively. For $25$th percentiles, $+5\%$ and $+0.4\%$ beyond MCCE and Focal loss, respectively. The TCE with $t=0.1$ seems to do better ($+1\%$ beyond Focal) which could be due to different sorting per model, and thus, different classes falling in those percentiles. In the ovrall metrics \cite{hashimoto2018fairness, li2019fair, mohri2019agnostic}, the worst-case IoU (among target classes) is $5\%$ and $+4\%$ better (higher) than MCCE and Focal loss, respectively. The standard deviation across classes is improved (decreased) by $+1\%$ comapred to MCCE and remains in the same regime as Focal loss.  Finally, qualitative results in Fig.~\ref{fig:comparison_1} further corroborate the impact of TCE in comparison with MCCE and Focal loss. The top two rows highlight improvement in ``rider'', ``bus'', and ``truck'', and the next two rows show improvement in ``tram'' and ``sidewalk'', most of which associate to the low-performing classes of the model trained with MCCE. There is plenty of room for extending this work. An avenue to explore is different implementations of TCE, especially (in a non-stochastic fashion) by directly plugging in \eqref{eq:term_m} and \eqref{eq:term_c} as the optimization loss function. In doing so, remedies have to be put in place to circumvent convergence issues. Further quantitative and qualitative results on ADE20k dataset are provided in the appendix. More experimentation on other datasets is left as future work.  \vspace{-0.3cm}

\section{Acknowledgment}
\label{sec:ack}
\vspace{-0.3cm}
The authors would like to thanks Shell Global Solutions International B.V. and Delft University of Technology (TU Delft) for their support and for the permission to publish this work. The authors extend their appreciation to Ahmad Beirami from Facebook AI for helpful discussions on tilted empirical risk minimization (TERM).

{\small
\bibliographystyle{ieeetr}
\bibliography{CVPR}

\begin{thebibliography}{10}

\bibitem{ronneberger2015u}
O.~Ronneberger, P.~Fischer, and T.~Brox, ``U-net: Convolutional networks for
  biomedical image segmentation,'' in {\em International Conference on Medical
  image computing and computer-assisted intervention}, pp.~234--241, Springer,
  2015.

\bibitem{rezaei2017conditional}
M.~Rezaei, K.~Harmuth, W.~Gierke, T.~Kellermeier, M.~Fischer, H.~Yang, and
  C.~Meinel, ``A conditional adversarial network for semantic segmentation of
  brain tumor,'' in {\em International MICCAI Brainlesion Workshop},
  pp.~241--252, Springer, 2017.

\bibitem{menze2015object}
M.~Menze and A.~Geiger, ``Object scene flow for autonomous vehicles,'' in {\em
  Proceedings of the IEEE conference on computer vision and pattern recognition
  (CVPR)}, pp.~3061--3070, 2015.

\bibitem{cordts2016cityscapes}
M.~Cordts, M.~Omran, S.~Ramos, T.~Rehfeld, M.~Enzweiler, R.~Benenson,
  U.~Franke, S.~Roth, and B.~Schiele, ``The cityscapes dataset for semantic
  urban scene understanding,'' in {\em Proceedings of the IEEE conference on
  computer vision and pattern recognition}, pp.~3213--3223, 2016.

\bibitem{volpi2015semantic}
M.~Volpi and V.~Ferrari, ``Semantic segmentation of urban scenes by learning
  local class interactions,'' in {\em Proceedings of the IEEE Conference on
  Computer Vision and Pattern Recognition (CVPR) Workshops}, pp.~1--9, 2015.

\bibitem{henry2018road}
C.~Henry, S.~M. Azimi, and N.~Merkle, ``Road segmentation in sar satellite
  images with deep fully convolutional neural networks,'' {\em IEEE Geoscience
  and Remote Sensing Letters}, vol.~15, no.~12, pp.~1867--1871, 2018.

\bibitem{geiger2013vision}
A.~Geiger, P.~Lenz, C.~Stiller, and R.~Urtasun, ``Vision meets robotics: The
  kitti dataset,'' {\em The International Journal of Robotics Research},
  vol.~32, no.~11, pp.~1231--1237, 2013.

\bibitem{shvets2018automatic}
A.~A. Shvets, A.~Rakhlin, A.~A. Kalinin, and V.~I. Iglovikov, ``Automatic
  instrument segmentation in robot-assisted surgery using deep learning,'' in
  {\em 2018 17th IEEE International Conference on Machine Learning and
  Applications (ICMLA)}, pp.~624--628, IEEE, 2018.

\bibitem{kass1988snakes}
M.~Kass, A.~Witkin, and D.~Terzopoulos, ``Snakes: Active contour models,'' {\em
  International journal of Computer Vision}, vol.~1, no.~4, pp.~321--331, 1988.

\bibitem{nock2004statistical}
R.~Nock and F.~Nielsen, ``Statistical region merging,'' {\em IEEE Transactions
  on pattern analysis and machine intelligence}, vol.~26, no.~11,
  pp.~1452--1458, 2004.

\bibitem{plath2009multi}
N.~Plath, M.~Toussaint, and S.~Nakajima, ``Multi-class image segmentation using
  conditional random fields and global classification,'' in {\em Proceedings of
  the 26th Annual International Conference on Machine Learning (ICML)},
  pp.~817--824, 2009.

\bibitem{minaee2019admm}
S.~Minaee and Y.~Wang, ``An {ADMM} approach to masked signal decomposition
  using subspace representation,'' {\em IEEE Transactions on Image Processing},
  vol.~28, no.~7, pp.~3192--3204, 2019.

\bibitem{long2015fully}
J.~Long, E.~Shelhamer, and T.~Darrell, ``Fully convolutional networks for
  semantic segmentation,'' in {\em Proceedings of the IEEE conference on
  computer vision and pattern recognition (CVPR)}, pp.~3431--3440, 2015.

\bibitem{zhao2017pyramid}
H.~Zhao, J.~Shi, X.~Qi, X.~Wang, and J.~Jia, ``Pyramid scene parsing network,''
  in {\em Proceedings of the IEEE conference on computer vision and pattern
  recognition (CVPR)}, pp.~2881--2890, 2017.

\bibitem{zhang2018context}
H.~Zhang, K.~Dana, J.~Shi, Z.~Zhang, X.~Wang, A.~Tyagi, and A.~Agrawal,
  ``Context encoding for semantic segmentation,'' in {\em Proceedings of the
  IEEE conference on Computer Vision and Pattern Recognition}, pp.~7151--7160,
  2018.

\bibitem{zhang2018exfuse}
Z.~Zhang, X.~Zhang, C.~Peng, X.~Xue, and J.~Sun, ``Exfuse: Enhancing feature
  fusion for semantic segmentation,'' in {\em Proceedings of the European
  Conference on Computer Vision (ECCV)}, pp.~269--284, 2018.

\bibitem{chen2018encoder}
L.-C. Chen, Y.~Zhu, G.~Papandreou, F.~Schroff, and H.~Adam, ``Encoder-decoder
  with atrous separable convolution for semantic image segmentation,'' in {\em
  Proceedings of the European conference on computer vision (ECCV)},
  pp.~801--818, 2018.

\bibitem{kirillov2019panoptic}
A.~Kirillov, K.~He, R.~Girshick, C.~Rother, and P.~Doll{\'a}r, ``Panoptic
  segmentation,'' in {\em Proceedings of the IEEE conference on computer vision
  and pattern recognition}, pp.~9404--9413, 2019.

\bibitem{cheng2019panoptic}
B.~Cheng, M.~D. Collins, Y.~Zhu, T.~Liu, T.~S. Huang, H.~Adam, and L.-C. Chen,
  ``Panoptic-deeplab,'' {\em arXiv preprint arXiv:1910.04751}, 2019.

\bibitem{wang2020deep}
J.~Wang, K.~Sun, T.~Cheng, B.~Jiang, C.~Deng, Y.~Zhao, D.~Liu, Y.~Mu, M.~Tan,
  X.~Wang, {\em et~al.}, ``Deep high-resolution representation learning for
  visual recognition,'' {\em IEEE transactions on pattern analysis and machine
  intelligence}, 2020.

\bibitem{jadon2020survey}
S.~Jadon, ``A survey of loss functions for semantic segmentation,'' in {\em
  2020 IEEE Conference on Computational Intelligence in Bioinformatics and
  Computational Biology (CIBCB)}, pp.~1--7, IEEE, 2020.

\bibitem{sudre2017generalised}
C.~H. Sudre, W.~Li, T.~Vercauteren, S.~Ourselin, and M.~J. Cardoso,
  ``Generalised dice overlap as a deep learning loss function for highly
  unbalanced segmentations,'' in {\em Deep learning in medical image analysis
  and multimodal learning for clinical decision support}, pp.~240--248,
  Springer, 2017.

\bibitem{ribera2018weighted}
J.~Ribera, D.~G{\"u}era, Y.~Chen, and E.~Delp, ``Weighted hausdorff distance: A
  loss function for object localization,'' {\em arXiv preprint
  arXiv:1806.07564}, vol.~2, 2018.

\bibitem{karimi2019reducing}
D.~Karimi and S.~E. Salcudean, ``Reducing the hausdorff distance in medical
  image segmentation with convolutional neural networks,'' {\em IEEE
  Transactions on medical imaging}, vol.~39, no.~2, pp.~499--513, 2019.

\bibitem{wong20183d}
K.~C. Wong, M.~Moradi, H.~Tang, and T.~Syeda-Mahmood, ``3d segmentation with
  exponential logarithmic loss for highly unbalanced object sizes,'' in {\em
  International Conference on Medical Image Computing and Computer-Assisted
  Intervention}, pp.~612--619, Springer, 2018.

\bibitem{pihur2007weighted}
V.~Pihur, S.~Datta, and S.~Datta, ``Weighted rank aggregation of cluster
  validation measures: a monte carlo cross-entropy approach,'' {\em
  Bioinformatics}, vol.~23, no.~13, pp.~1607--1615, 2007.

\bibitem{xie2015holistically}
S.~Xie and Z.~Tu, ``Holistically-nested edge detection,'' in {\em Proceedings
  of the IEEE international conference on computer vision}, pp.~1395--1403,
  2015.

\bibitem{lin2017focal}
T.-Y. Lin, P.~Goyal, R.~Girshick, K.~He, and P.~Doll{\'a}r, ``Focal loss for
  dense object detection,'' in {\em Proceedings of the IEEE international
  conference on computer vision}, pp.~2980--2988, 2017.

\bibitem{hashimoto2018fairness}
T.~Hashimoto, M.~Srivastava, H.~Namkoong, and P.~Liang, ``Fairness without
  demographics in repeated loss minimization,'' in {\em International
  Conference on Machine Learning}, pp.~1929--1938, PMLR, 2018.

\bibitem{duchi2016variance}
J.~Duchi and H.~Namkoong, ``Variance-based regularization with convex
  objectives,'' {\em arXiv preprint arXiv:1610.02581}, 2016.

\bibitem{li2020tilted}
T.~Li, A.~Beirami, M.~Sanjabi, and V.~Smith, ``Tilted empirical risk
  minimization,'' {\em arXiv preprint arXiv:2007.01162}, 2020.

\bibitem{xiao2018unified}
T.~Xiao, Y.~Liu, B.~Zhou, Y.~Jiang, and J.~Sun, ``Unified perceptual parsing
  for scene understanding,'' in {\em Proceedings of the European Conference on
  Computer Vision (ECCV)}, pp.~418--434, 2018.

\bibitem{zhou2019semantic}
B.~Zhou, H.~Zhao, X.~Puig, T.~Xiao, S.~Fidler, A.~Barriuso, and A.~Torralba,
  ``Semantic understanding of scenes through the ade20k dataset,'' {\em
  International Journal of Computer Vision}, vol.~127, no.~3, pp.~302--321,
  2019.

\bibitem{li2019fair}
T.~Li, M.~Sanjabi, A.~Beirami, and V.~Smith, ``Fair resource allocation in
  federated learning,'' {\em arXiv preprint arXiv:1905.10497}, 2019.

\bibitem{mohri2019agnostic}
M.~Mohri, G.~Sivek, and A.~T. Suresh, ``Agnostic federated learning,'' {\em
  arXiv preprint arXiv:1902.00146}, 2019.

\end{thebibliography}
}

\clearpage
\appendix

\onecolumn

\section{More Evaluation Results}
\label{sec:more_eval_res}
\vspace{-0.3cm}
%
\begin{table*}[th!]
    \footnotesize
	\caption{Performance comparison on ADE20k \emph{validation} set, bottom 22 classes of MCCE.}
	\vspace{-0.35cm}
	\label{tb:ade20k_bottom_val}
	\centering
	{\tabcolsep=0pt\def\arraystretch{1.0}
	\begin{tabularx}{400pt}{@{}l Y Y Y Y Y Y Y Y Y Y Y Y@{}}
		\toprule
		Method     	& land & lake & shower & blanket & step & hill & bag & crt scrn. & tray & stage & hovel & dirt track \\
		\midrule
		MCCE \cite{xiao2018unified}		 & 0.00 & 1.96 & {\bf2.62} & 2.95 & 5.47 & {\bf5.52} & 5.81 & 6.12 & 6.94 & {\bf7.66} & 8.77 & {\bf10.37} \\
		\rowcolor{LightCyan}
		TCE \tiny{$t=.1$}		  & {\bf3.58} & {\bf34.94} & 1.44 & {\bf7.13} & {\bf7.54} & 3.46 & 6.04 & {\bf8.50} & 5.58 & 6.82 & {\bf9.14} & 1.71\\
		\rowcolor{LightYellow}
		TCE \tiny{$t=1$}		  & 1.08 & 14.64 & 2.30 & 4.60 & 3.19 & 4.76 & {\bf6.19} & 0.86 & {\bf9.18} & 7.37 & 8.49 & 2.04\\
		\toprule
		\toprule
		continued	 & truck & river & bannister & canopy & glass & fountain & barrel & box & pole	&tower &\multicolumn{1}{c}{\cellcolor{gray!35}mIoU} & \multicolumn{1}{c}{\cellcolor{gray!35}mIoU\scriptsize{(all)}}\\
		\midrule
        MCCE \cite{xiao2018unified}  & 10.70 & {\bf10.71} & 11.55 & {\bf11.62} & 12.63 & 12.67 & 17.68 & {\bf18.57} & 19.37 & 19.48 &\multicolumn{1}{c}{\cellcolor{gray!35}9.51}  &\multicolumn{1}{c}{\cellcolor{gray!35}\bf43.87} \\
		\rowcolor{LightCyan}
		TCE \tiny{$t=.1$}	 & 12.85 & 2.88 & 12.27 & 9.37 & {12.83} & {\bf18.92} & {f30.93} & 10.58 & 16.53 & 19.98 &\multicolumn{1}{c}{\cellcolor{gray!35}\bf11.56}  &\multicolumn{1}{c}{\cellcolor{gray!35}\cellcolor{gray!35}41.32} \\
		\rowcolor{LightYellow}
		TCE \tiny{$t=1$}	   & {\bf17.17} & 10.49 & {\bf12.78} & 10.51 & {\bf13.34} & 15.87 & {\bf45.20} & 17.36 & {\bf19.61} & 18.74 & \multicolumn{1}{c}{\cellcolor{gray!35}11.17} & \multicolumn{1}{c}{\cellcolor{gray!35}41.77} \\
		\bottomrule
	\end{tabularx}}
	\vspace{-0.1cm}
\end{table*}
%
%
\begin{table*}[th!]
    \footnotesize
	\caption{Performance comparison on ADE20k \emph{validation} set, top 22 classes of MCCE.}
	\vspace{-0.35cm}
	\label{tb:ade20k_top_val}
	\centering
	{\tabcolsep=0pt\def\arraystretch{1.0}
	\begin{tabularx}{400pt}{@{}l Y Y Y Y Y Y Y Y Y Y Y Y@{}}
		\toprule
		Method     	& sky & p. table & bed & toilet & road & ceiling & car & building & tent & bus & floor & person \\
		\midrule
		MCCE \cite{xiao2018unified}		  & 	{\bf93.92} & {\bf91.99} & {\bf86.39} & 83.02 & {\bf82.70} & {81.76} & {81.66} & {\bf81.41} & 80.50 & 79.12 & {\bf78.91} & {\bf78.42} \\
		\rowcolor{LightCyan}
		TCE \tiny{$t=.1$}		  & 93.81 & 91.79 & 85.80 & {84.58} & 80.78 & 81.40 & 80.69 & 79.31 & 84.76 & 79.76 & 77.82 & 76.64 \\
		\rowcolor{LightYellow}
		TCE \tiny{$t=1$}	   & {\bf93.97} & 90.27 & 85.33 & {\bf84.78} & 80.97 & {\bf81.94} & {\bf81.85} & 80.18 & {\bf91.93} & {\bf83.64} & 77.47 & 77.27 \\

		\toprule
		\toprule
		continued	 & stove & wall & cradle & bathtub & tree & painting & dish w. & refrig. & runway & sink &\multicolumn{1}{c}{\cellcolor{gray!35}mIoU} & \multicolumn{1}{c}{\cellcolor{gray!35}mIoU\scriptsize{(all)}}\\
		\midrule
		MCCE \cite{xiao2018unified}	 & {\bf76.14} & {\bf74.68} & 73.67 & {72.66} & {\bf71.14} & {\bf70.96} & {\bf70.70} & {70.62} &{\bf70.48} &69.56 &\multicolumn{1}{c}{\cellcolor{gray!35}\bf78.20} &\multicolumn{1}{c}{\cellcolor{gray!35}\bf43.87} \\
		\rowcolor{LightCyan}
		TCE \tiny{$t=.1$}		  & 68.66 & 74.08 &{\bf80.17} & 70.95 & 70.01 & 64.60 & 67.89 & 68.46 &62.61 &69.65	&\multicolumn{1}{c}{\cellcolor{gray!35}77.25} &\multicolumn{1}{c}{\cellcolor{gray!35}41.32} \\
		\rowcolor{LightYellow}
		TCE \tiny{$t=1$}	   &  70.84 & 73.92 & 77.89 & {\bf75.39} & 70.59 & 66.24 & 64.31 & {\bf74.19} & 66.98 & {\bf69.70} & \multicolumn{1}{c}{\cellcolor{gray!35}{\bf78.17}} & \multicolumn{1}{c}{\cellcolor{gray!35}41.77} \\
		\bottomrule
	\end{tabularx}}
	\vspace{-0.1cm}
\end{table*}
%
%
	
\begin{table*}[th!]
    \footnotesize
	\caption{Fairness criteria for ADE20k \cite{zhou2019semantic}.}
	\vspace{-0.35cm}
	\label{tb:fairness_ADE20k}
	\centering
	{\tabcolsep=0pt\def\arraystretch{1.0}
	
		\begin{tabularx}{230pt}{l  s  s | m m | s s }
			\toprule
			& \multicolumn{2}{c}{sorted $15\%$ } & \multicolumn{2}{c}{$(15^{th} \textup{perc., mIoU})$}  & \multicolumn{2}{c}{overall}
		    \tabularnewline
			\cmidrule(lr){2-3}\cmidrule(lr){4-5}\cmidrule(lr){6-7}
			Method   & bottom     & top        & bottom    &top  & worst & std. 				\tabularnewline 
			\midrule
			MCCE \cite{chen2018encoder}  & 9.51 &{78.20}& (19.55, {\bf9.51}) &(69.14, 78.20)  & 0.0  & {\bf21.95} \\
			\rowcolor{LightCyan}
			TCE \tiny{$t=.1$} & {\bf11.56} &77.25 &(15.31, 8.04) &(67.95, 77.62) & {\bf1.44}  & 22.39 \\
			\rowcolor{LightYellow}
			TCE \tiny{$t=1$} & {11.17} &78.17 &(15.67, 7.99) &(67.26, {78.49}) & 0.86  & 22.79 \\
			\bottomrule
	\end{tabularx}}
\end{table*} 

We further assessed the impact of the proposed tilted cross-entropy (TCE) on yet another commonly adopted datasets for semantic segmentation; i.e., ADE20k \cite{zhou2019semantic}. ADE20k contains $20,100$ images for training and $2,000$ for validations from $150$ object and stuff classes. As such, compared to Cityscapes \cite{cordts2016cityscapes}, the typical scenes in ADE20k can be more complex in that they can potentially contain more target classes per image. For experiments on ADE20k, we have used the UPerNet \cite{xiao2018unified} with ResNet-101 backbone as our reference implementation of multi-class cross-entropy (MCCE) and on top that we have implemented the proposed TCE. UPerNet is among the top performing model architectures for ADE20k. Following \cite{xiao2018unified}, we used minibatch SGD with learning rate $l_r = 0.01$ and momentum $0.9$ for all models, and adjusted for a total minibatch size of $8$. The reported results of \cite{xiao2018unified} are based on our own trainings, for the sake of a fair comparison. Image crop size and other pre/post-processing parameters are set per default as suggested in \cite{xiao2018unified}. Our evaluation strategy is exactly the same as explained for Cityspaces in Section~\ref{sec:exp} except that here we consider sorted $15\%$ (bottom and top 22 classes) and bottom and top $15$th percentile. This is because ADE20k contains much larger number of classes compared to Cityscapes ($150$ vs $19$).  

\begin{figure*}[t]
	\centering

    \includegraphics[trim={0cm 3.88cm 0cm 0cm},clip,width=0.9\textwidth]{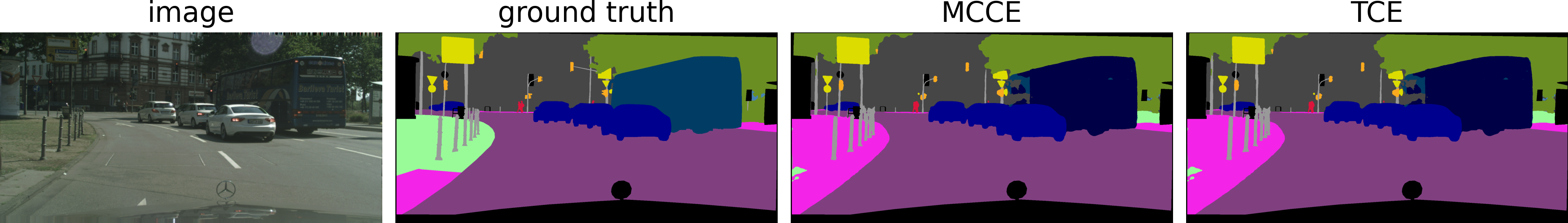}
	\begin{tikzonimage}[trim={0cm 0.35cm 0cm 0.55cm},clip,width=0.9\textwidth]{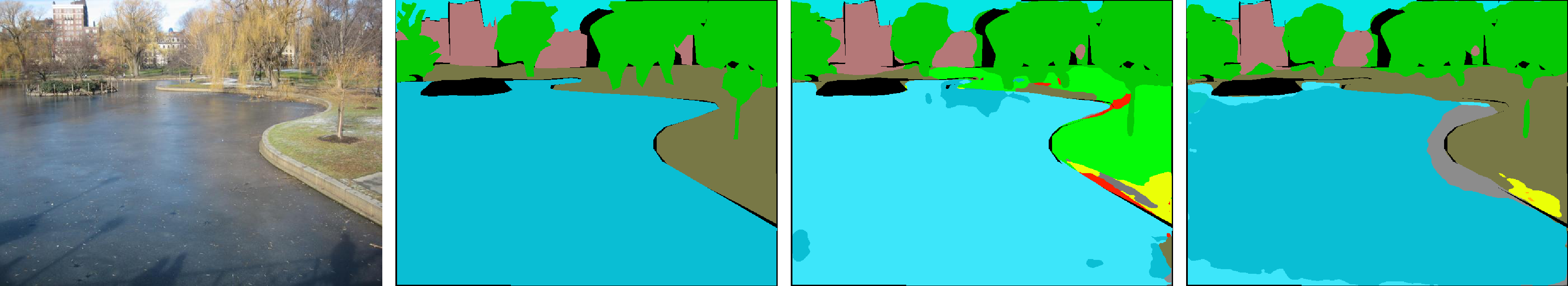}
		\draw[yellow, very thick, rounded corners, dashed] (0.755,0.01) rectangle (0.969,0.77);
		\draw[yellow, very thick, rounded corners, dashed] (0.9,0.2) rectangle (0.999,0.88);
	\end{tikzonimage}
	\begin{tikzonimage}[width=0.9\textwidth]{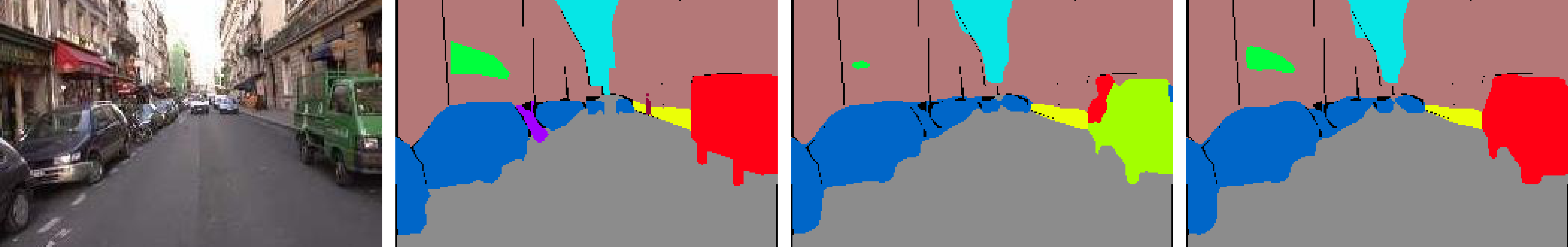}
		\draw[yellow, very thick, rounded corners, dashed] (0.94,0.225) rectangle (0.999,0.785);
	\end{tikzonimage}
	\begin{tikzonimage}[width=0.9\textwidth]{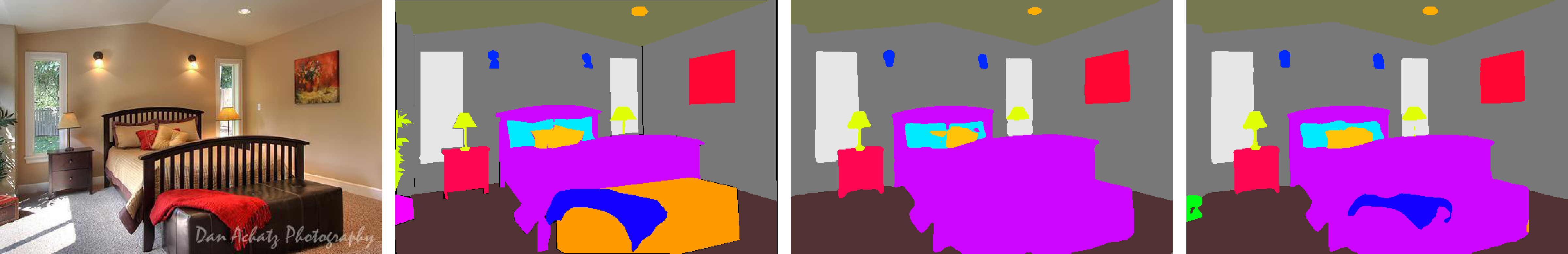}
		\draw[yellow, very thick, rounded corners, dashed] (0.85,0.1) rectangle (0.93,0.337);
	\end{tikzonimage}
	\begin{tikzonimage}[width=0.9\textwidth]{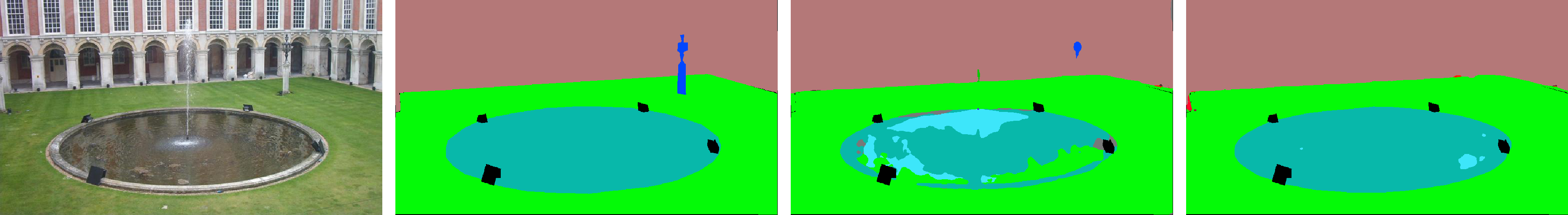}
	    \draw[yellow, very thick, rounded corners, dashed] (0.78,0.07) rectangle (0.969,0.58);
	\end{tikzonimage}
	\begin{tikzonimage}[width=0.9\textwidth]{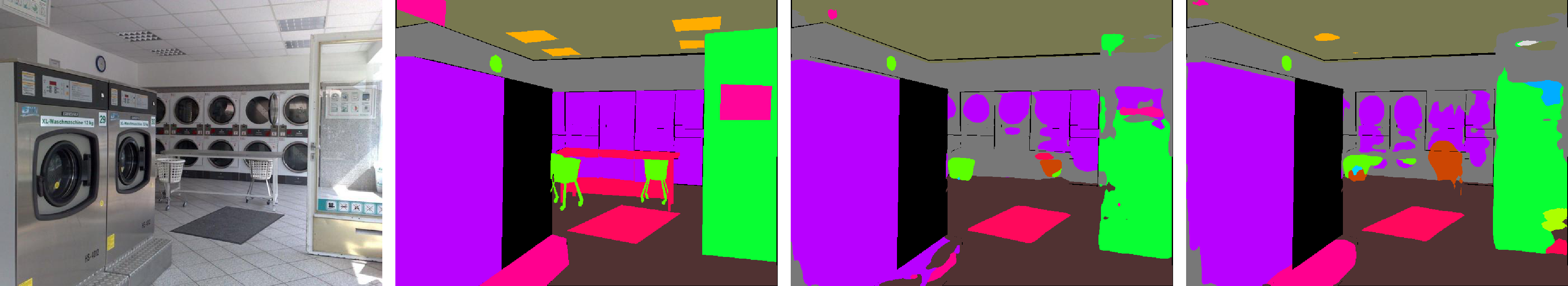}
		\draw[yellow, very thick, rounded corners, dashed] (0.8,0.01) rectangle (0.87,0.20);
	\end{tikzonimage}
	\caption{Impact of TCE on improving low-performing classes of MCCE. Best view in color with $300\%$ zoom. From the top row to bottom, tilted cross-entropy (TCE) is providing a more consistent label map compared to standard multi-class cross-entropy (MCCE) for ``lake'', ``truck'', ``blanket'', ``fountain'' and ``step'' all which lie within the low-performing bottom $15\%$ ($22$) classes of ADE20k.}
	\label{fig:comparison_2}
\end{figure*}

Table~\ref{tb:ade20k_bottom_val}, compares the sorted mIoU breakdown of MCCE (UperNet \cite{xiao2018unified}) for its bottom $15\%$ ($22$) classes against the same model trained with TCE. Here, again we see improvement in the low-performing classes, which is also reflected in the mIoU of these $22$ classes (for both $t = 0.1$ and $1$). Conversely, the overall mIoU (denoted as mIoU(all)) has dropped for ADE20k irrespective of the choice of the tilting parameter $t$. To reiterate, TCE favors a less varied (and more fair) performance across classes, and not an improved overall mIoU. Table~\ref{tb:ade20k_top_val} provides the top part. As can be seen, here for most classes, MCCE outperforms TCE which confirms that the improvement in bottom (low-performing) classes is coming at the cost of performance degradation in top performing classes. The complete MIoU breakdown of ADE20k with $150$ classes would not fit in two tables. Instead Table~\ref{tb:fairness_ADE20k} summarizes the fairness measures for ADE20k. Here, the trend is less consistent compared to Cityscapes. TCE shows improvement in sorted percentage and overall fairness measures, but in percentile analysis this is not visible. This could be because typical ADE20k images contain more target classes. As such, in Algorithm~\ref{alg:sotchastc_term}, every time we samples from $\mathcal{D}^t_c$ to improve class $c$, we also involve several other classes. A possible remedy could be to tilt at class level per image. Finally, Figure~\ref{fig:comparison_2} provides further qualitative examples illustrating the impact of TCE on ADE20k dataset. From the top row to bottom, TCE is providing a more consistent label map compared to MCCE for ``lake'', ``truck'', ``blanket'', ``fountain'' and ``step'' all which lie within the low-performing bottom $15\%$ ($22$) classes of ADE20k.

\end{document}